\documentclass{article}

\PassOptionsToPackage{numbers, compress}{natbib}

\usepackage[preprint]{neurips_2026}

\usepackage[utf8]{inputenc} 
\usepackage[T1]{fontenc}    
\usepackage{hyperref}       
\usepackage{url}            
\usepackage{booktabs}       
\usepackage{amsfonts}       
\usepackage{nicefrac}       
\usepackage{microtype}      
\usepackage{xcolor}         
\usepackage{pifont}
\usepackage{graphicx}
\usepackage{algorithm}
\usepackage{algpseudocode}
\usepackage{amsmath}     
\usepackage{amssymb}
\newcommand{\cmark}{\ding{51}}
\newcommand{\xmark}{\ding{55}}

\title{CR-Refiner: An Object-Centric Optimal Transport Reranker for Edit-Conditioned 3D Scene Retrieval}

\author{%
  Hao Wu$^{1,2}$ \quad
  Jinjing Zhu$^{2}$ \quad
  Nanyu Wu$^{2}$ \quad
  Qianyi Cai$^{2}$ \quad
  Heyi Lin$^{2}$ \quad
  Hao Wang$^{2}$ \quad
  Hui Xiong$^{1,2}$\thanks{Corresponding author.} \\[6pt]
  $^{1}$Hong Kong University of Science and Technology \\
  $^{2}$Hong Kong University of Science and Technology (Guangzhou) \\
}

\begin{document}

\maketitle

\begin{abstract}

Edit-conditioned 3D scene retrieval pairs a reference 3D room with a natural-language modification and retrieves rooms from a corpus that satisfy the edit. Three lines of prior work each fall short on this task. 2D composed image retrieval reasons over pixel-level edits and has no primitive for 3D object sets. 3D foundation encoders embed individual objects but cannot compose at the scene level. 3D scene-grounding methods localize references inside a static scene rather than rank modified rooms across a corpus. We present \textbf{CR-Refiner}, a training-free reranker that wraps any base retriever's top-$K$ candidates with three components. A frozen LLM parses the edit into a structured query entity, and each candidate is scored by an unbalanced optimal-transport problem over a $1{\times}G$ cost matrix coupling category, style, material, and geometry. The unbalanced solver lets the single-entity query drop mass on irrelevant objects, modelling the asymmetry directly. An axis-conditional structural prior adds size-keyword cues for geometric edits and subject-anchor direction cues for spatial edits. An LLM verifier refines the top three candidates with continuous confidence. Because no benchmark evaluates compositional matching over 3D object sets, we additionally release \textbf{3D-CER}, $4{,}963$ edit-conditioned queries over a $23{,}381$-room indoor corpus across five edit axes, with multi-positive ground truth, CIRR-style hard subsets, and zero-target adversarials. Across three qualitatively distinct base retrievers, CR-Refiner consistently improves hard-subset R@$1$ and mAP@$10$ on every edit axis.

\end{abstract}

\section{Introduction}
\label{sec:intro}

3D scene retrieval is widely used in interior design, embodied agents, and retrieval-augmented scene synthesis. A user submits either a text description matched against scene captions and graph metadata~\citep{chen2020scanrefer,zhang2023multi3drefer}, or a reference scene matched by global appearance. Neither interface fits a common workflow. A designer reviewing a draft room often wants to filter the corpus by an \textit{edit} applied to the current scene, such as ``swap the king-size bed for a single bed''.

We study \textit{edit-conditioned 3D scene retrieval}, the task of retrieving rooms that satisfy a natural-language modification of a given reference room. The reference room is structured. Each object carries a category, a style, a material, and a 3D bounding box. The edit instruction names one object and specifies a target attribute or a relative position. Following the protocol of 2D composed image retrieval~\citep{liu2021image,baldrati2023zero}, performance is measured by Recall@$K$ and mAP@$10$ over a CIRR-style hard-distractor subset.

The matching problem is asymmetric. A candidate room contains many objects but the edit names only one. The retriever must locate the edited slot, check its attributes, and verify that the rest of the room is preserved. Three lines of prior work each address only part of this problem. 2D composed image retrieval~\citep{vo2018composingtextimageimage,liu2021image,baldrati2022effective,baldrati2023zero,karthik2023vision,saito2023pic2word,gu2024compodiff,gu2024languageonlyefficienttrainingzeroshot} reasons over pixel-level edits and has no primitive for 3D object sets. 3D foundation encoders~\citep{zhang2021pointclippointcloudunderstanding,liu2023openshape,zhou2023uni3d,xue2024ulip} embed individual objects but cannot compose at the scene level. 3D scene-grounding methods~\citep{chen2020scanrefer,zhang2023multi3drefer,zhu20233dvistapretrainedtransformer3d,hong20233dllminjecting3dworld,xu2024pointllmempoweringlargelanguage,miao2024scenegraphloccrossmodalcoarsevisual} localize references inside a static scene rather than retrieve modified ones across a corpus.

We present \textbf{CR-Refiner}, a training-free reranker that wraps any base retriever's top-$K$ candidates with three components. The \textit{coupled-OT backbone} scores each candidate by an unbalanced optimal-transport problem over a $1{\times}G$ cost matrix coupling category, style, material, and geometry, where the unbalanced solver lets the single-entity query drop mass on irrelevant objects rather than splitting unit mass across $G$ candidate objects. An axis-conditional \textit{structural prior} adds bounding-box size cues for geometric edits and subject-anchor direction cues for spatial edits, with the router reusing the LLM-parsed axis at no additional cost. An \textit{LLM verifier} refines the top-three candidates with continuous confidence to disambiguate near-ties. Section~\ref{sec:method} details each component.

Validating CR-Refiner requires a benchmark with multi-axis edits, multi-positive ground truth, and hard-negative subsets that distinguish structural matching from caption matching. Because no such resource existed, we release \textbf{3D-CER}: edit-conditioned queries over a $23{,}381$-room 3D-FRONT/3D-FUTURE corpus across five edit axes (style, material, spatial, geometric, compound), with multi-positive ground truth, CIRR-style 20-distractor hard subsets, and zero-target adversarials. The benchmark is built end-to-end with no human annotation; correctness is enforced by programmatic predicates and an independent cross-model audit. Across three qualitatively distinct base retrievers, CR-Refiner consistently improves hard-subset R@$1$ over the strongest baseline on every edit axis (Section~\ref{sec:exp:main}).

In summary, this paper makes the following contributions:
\begin{itemize}
    \item We propose \textbf{CR-Refiner}, the first training-free reranker for edit-conditioned 3D scene retrieval, built on object-centric coupled optimal transport with an axis-conditional structural prior and an LLM verifier.
    \item We release \textbf{3D-CER}, a benchmark of $4{,}963$ edit-conditioned queries over $23{,}381$ indoor rooms across five edit axes, with multi-positive ground truth, CIRR-style hard subsets, and zero-target adversarials.
    \item We empirically validate retriever-agnostic plug-in compatibility on three qualitatively distinct base retrievers, and decompose the candidate-generation versus reranking regimes on the full $23$K-room corpus.
\end{itemize}

\section{Related Work}
\label{sec:related}

\paragraph{Composed image retrieval (CIR).}
The CIR paradigm, where a reference image and an edit text are composed into a target-image query, was established by CIRR~\citep{liu2021image}, which introduced the subset-based hard-negative protocol that we adopt for our hard subsets. CIRCO~\citep{baldrati2023zero} reformulated CIR as multi-positive retrieval to address CIRR's false-negative issue, an approach we follow with mAP@$10$ as the primary metric. Among training-free CIR methods, CIReVL \citep{karthik2023vision} first captions the reference image with a VLM and then prompts an LLM to rewrite the caption together with the edit into a target description, while Pic2Word~\citep{saito2023pic2word} maps the reference image to a pseudo-word token in CLIP's token embedding space and concatenates it with the edit text. LinCIR~\citep{gu2024languageonlyefficienttrainingzeroshot} extends this textual-inversion paradigm with a language-only training strategy for better backbone scalability, and CompoDiff~\citep{gu2024compodiff} models CIR as conditional latent-diffusion editing on CLIP embeddings. The shared limitation across this line is the 2D pixel modality: none of these methods can reason over 3D object sets, bounding-box geometry, or scene-level composition. We adapt CIRR's subset protocol and CIRCO's multi-positive evaluation to a new modality, and use CIReVL and Pic2Word as base-retriever baselines (Section~\ref{sec:exp:plugin}) to test plug-in compatibility.

\paragraph{3D scene language datasets.}
ScanRefer~\citep{chen2020scanrefer} and the ReferIt3D~\citep{10.1007/978-3-030-58452-8_25} family annotate referring expressions over static 3D scenes; Multi3DRefer~\citep{zhang2023multi3drefer} generalizes to zero, one, and multi-target grounding, whose adversarial design we reuse for our zero-target queries; ScanQA~\citep{azuma2022scanqa3dquestionanswering} addresses 3D question answering; and SceneVerse~\citep{jia2024sceneverse} scales 3D-language pretraining over $68$K scenes. 3D-LLM\citep{hong20233dllminjecting3dworld} brings 3D scenes and PointLLM\citep{xu2024pointllmempoweringlargelanguage} brings 3D objects into instruction-tuned multimodal LLMs for free-form reasoning. Closer to retrieval, SceneGraphLoc~\citep{miao2024scenegraphloccrossmodalcoarsevisual} matches a query image against a database of compact 3D scene graphs for cross-modal coarse localization. The shared limitation is the static-scene assumption: queries describe what is in a scene, not what an edit would produce. To our knowledge, 3D-CER is the first benchmark that composes a reference scene with a natural-language modification and requires retrieving the post-edit target.

\paragraph{3D foundation encoders and edit-conditioned 3D generation.}
OpenShape~\citep{liu2023openshape}, Uni3D~\citep{zhou2023uni3d}, and ULIP-2~\citep{xue2024ulip} produce CLIP-aligned single-object embeddings, but offer no scene-level composition primitive. ChangeIt3D~\citep{achlioptas2023shapetalk}, built on the ShapeTalk dataset, addresses language-conditioned per-shape editing rather than scene-level retrieval. For scene-level generation, DiffuScene~\citep{tang2024diffuscenedenoisingdiffusionmodels} and InstructScene~\citep{lin2024instructsceneinstructiondriven3dindoor} synthesize indoor 3D layouts via diffusion (unconditional and instruction-driven, respectively), and EditRoom~\citep{zheng2025editroomllmparameterizedgraphdiffusion} generates edited rooms through diffusion rather than retrieving from a fixed corpus. The shared limitation is task formulation: each treats edits as a generation problem on a single object or scene rather than retrieval over a corpus. We treat the corpus as fixed and frame the problem as reranking, which permits zero-cost deployment on any indoor 3D collection without per-scene generation.

\section{CR-Refiner}
\label{sec:method}



\subsection{Preliminaries: Query Entity and Candidate Specification}
\label{sec:method:prelim}

We denote the reference 3D room as $R$ and the natural-language edit text as $t$. A frozen LLM parses $(R, t)$ into a structured query entity $q = (c_q, s_q, m_q, d_q, \mathrm{axis})$ with five fields: target category $c_q$, target style $s_q$, target material $m_q$, optional spatial direction $d_q \in \{\text{left, right, front, behind, near, far}\}$, and the predicted edit axis $\mathrm{axis} \in \{\text{STY, MAT, SPA, GEO, CMP}\}$ (style, material, spatial, geometric, compound). The base retriever returns a candidate pool $\mathcal{P}$ of size $K$. Each candidate room $C \in \mathcal{P}$ comes with a structured spec listing its $G$ objects $\{o_j\}_{j=1}^{G}$, where each object $o_j = (c_j, s_j, m_j, p_j)$ has a category, optional style, optional material, and a 3D bounding-box centre $p_j \in \mathbb{R}^3$. All attribute fields use the 3D-FUTURE attribute vocabulary; missing values are preserved as nulls rather than imputed. CR-Refiner produces a final score $\hat{s}(C)$ for each candidate, and the reranked output sorts $\mathcal{P}$ by $\hat{s}$ in ascending order.

\subsection{Coupled-OT Backbone}
\label{sec:method:backbone}

\paragraph{Motivation.}
An edit instruction names \textit{one} subject (e.g.,\ ``the king-size bed'') while a candidate room contains \textit{many} objects, most of which are unrelated to the edit. A plain dot-product over object embeddings cannot exploit this one-to-many asymmetry: it treats every object as equally salient and pollutes the score with noise from unmentioned furniture. We therefore frame reranking as a transport problem over an asymmetric $1{\times}G$ cost matrix.

\paragraph{Cost matrix.}
For each candidate object $o_j$, we define four hard mismatch indicators that compare $q$ against $o_j$ symbol by symbol. We use the index set $\mathcal{A} = \{\mathrm{cat}, \mathrm{sty}, \mathrm{mat}, \mathrm{dir}\}$ to label these four channels; these lowercase tags are attribute-channel names internal to the cost matrix and are distinct from the uppercase edit-axis labels $\{\text{STY, MAT, SPA, GEO, CMP}\}$ introduced in \S\ref{sec:method:prelim}. The category indicator $\ell_{\mathrm{cat}}(q, o_j) \in \{0, 1\}$ is $0$ iff $c_q$ and $c_j$ are string-equal. The style indicator $\ell_{\mathrm{sty}}(q, o_j) \in \{0, 0.5, 1\}$ takes value $0$ when $s_q$ matches the candidate's style (either $o_j$'s own style $s_j$ or any style in the candidate room's style multiset, as a room-level fallback for missing object-level annotation), $1$ when both $s_j$ and the room style are present but neither matches, and $0.5$ when both are missing. The material indicator $\ell_{\mathrm{mat}}(q, o_j) \in \{0, 0.5, 1\}$ follows the same scheme without the room-level fallback, since material is an object-level attribute. The direction indicator $\ell_{\mathrm{dir}}(q, o_j) \in \{0, 1\}$ is $0$ iff the sign of the bounding-box centre coordinate $p_j$ along the relevant axis matches the parsed direction $d_q$. The four indicators aggregate into a single mismatch cost
\begin{equation}
M_{1,j} = \frac{\sum_{k \in \mathcal{A}} w_{k}\,\ell_{k}(q, o_j)}{\sum_{k \in \mathcal{A}} w_{k}},
\label{eq:costmatrix}
\end{equation}
with weights $w_{\mathrm{cat}}{=}0.4$, $w_{\mathrm{sty}}{=}0.3$, $w_{\mathrm{mat}}{=}0.2$, and $w_{\mathrm{dir}}{=}0.1$. A term contributes only if the corresponding query field is non-null, so a geometric query (no $s_q$, no $m_q$) is not penalised by missing style or material. The intentionally low direction weight reflects that bounding-box centres in 3D-FRONT live in scene-specific world frames, making $\ell_{\mathrm{dir}}$ unreliable on its own; the structural prior (\S\ref{sec:method:prior}) restores this signal on spatial queries.

\begin{figure}[t]
  \centering
  \includegraphics[width=\linewidth]{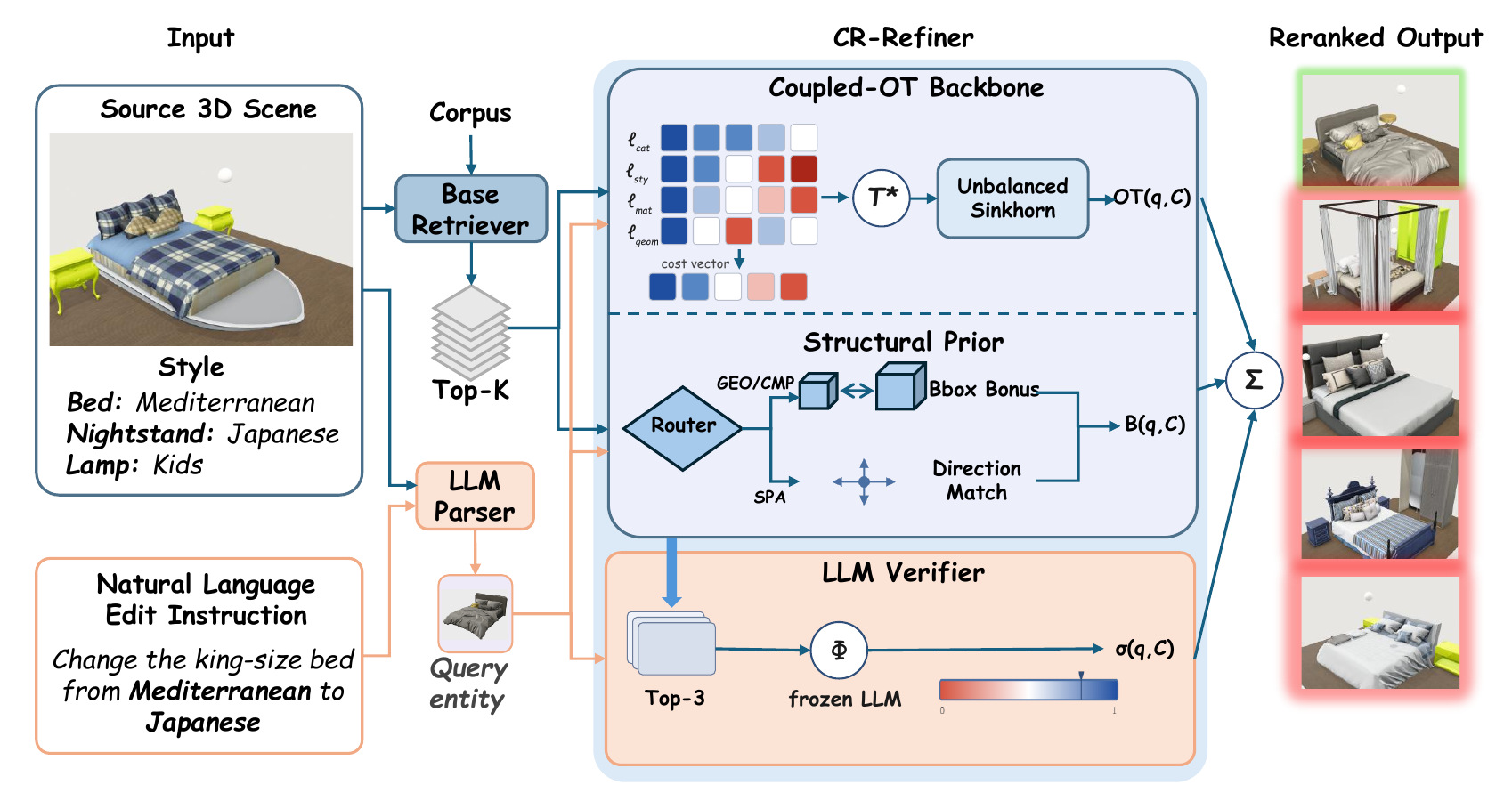}
  \vspace{-13pt}
  \caption{\textbf{Overview of CR-Refiner.} Given a reference 3D room $R$ and a natural-language edit $t$, a base retriever returns a top-$K$ candidate pool $\mathcal{P}$ while a frozen LLM parses $(R, t)$ into a structured query entity $q$. CR-Refiner rescores each candidate via a \textbf{Coupled-OT Backbone}, an axis-conditional \textbf{Structural Prior}, and an \textbf{LLM Verifier} applied to the top-$3$ candidates after the backbone and prior; the three signals are combined additively into the final reranking score (Eq.~\ref{eq:final}).}
  \vspace{-10pt}
  \label{fig:pipeline}
\end{figure}

\paragraph{Unbalanced Sinkhorn solver.}
Stacking the four indicators across all $G$ candidate objects yields the cost matrix $\mathbf{M} \in \mathbb{R}^{1 \times G}$ with entries $M_{1,j}$ from Eq.~\ref{eq:costmatrix}. With the asymmetric source distribution $a = [1] \in \mathbb{R}^{1}$ and uniform target $b = \mathbf{1}_{G}/G$, we solve the regularised unbalanced transport problem~\citep{chizat2018scaling}
\begin{equation}
\mathbf{T}^{\star} = \arg\min_{\mathbf{T} \geq 0}\; \langle \mathbf{T}, \mathbf{M} \rangle - \varepsilon\,H(\mathbf{T}) + \tau\,\mathrm{KL}(\mathbf{T}\mathbf{1}_{G}\,\|\,a),
\label{eq:unbalanced}
\end{equation}
where $\mathbf{T} \in \mathbb{R}_{\geq 0}^{1 \times G}$ is the transport plan, $\langle \mathbf{T}, \mathbf{M} \rangle = \sum_{i,j} T_{ij} M_{ij}$ is the Frobenius inner product, $H(\mathbf{T}) = -\sum_{i,j} T_{ij}\log T_{ij}$ is the Shannon entropy of $\mathbf{T}$, $\mathbf{T}\mathbf{1}_{G} \in \mathbb{R}$ is the row marginal of $\mathbf{T}$, $\varepsilon$ is the entropic regularisation weight, and $\tau$ is the KL slack on the row marginal that allows the solver to drop mass on rows with no plausible match instead of forcing the unit query mass to be split across all $G$ candidate objects. We use $\varepsilon{=}0.01$ and $\tau{=}1.0$, both selected on the validation split, and take the transport cost $\mathrm{OT}(q, C) = \langle \mathbf{T}^{\star}, \mathbf{M} \rangle$ as the backbone score.


\subsection{Structural Prior}
\label{sec:method:prior}

\paragraph{Motivation.}
The backbone consumes only categorical, stylistic, material, and sign-of-coordinate signals. Two channels remain invisible. First, fine-grained bounding-box geometry: the symbolic $\ell_{\mathrm{dir}}$ is deliberately weak, and size-keyword cues such as ``smaller'' or ``king-size'' are not used at all. Second, inter-object adjacency: spatial edits like ``move the chair to the right of the desk'' depend on relations between pairs of objects, not on any single object's position. These two channels call for different primitives. Size cues are localised to a single object and read directly from the category vocabulary. Direction cues require comparing two objects in the same room, which we handle with a subject-anchor scheme. We therefore design an axis-conditional structural prior that selects the right primitive for each edit type, using the parser-predicted $\mathrm{axis}$ at no additional cost.

\paragraph{Size-class bonus (geometric and compound branch).}
For queries with $\mathrm{axis} \in \{\text{GEO, CMP}\}$, we compute a size-class bonus that detects size modifiers (\textit{small / single / compact / smaller} for small; \textit{king / large / larger / big} for large) from the parsed paraphrase. The bonus rewards candidates that contain at least one object in the queried category family at the matching size class. We apply a small penalty when the family is present at the wrong size class. STY and MAT queries receive zero bonus from this branch and rely entirely on the OT backbone's $\ell_{\mathrm{sty}}$ and $\ell_{\mathrm{mat}}$ terms.

\paragraph{Direction bonus (spatial and compound branch).}
For queries with $\mathrm{axis} \in \{\text{SPA, CMP}\}$, we compute a subject-anchor direction bonus. We identify the subject object in the candidate by category match against $c_q$, and pick an anchor as the highest-mention non-subject category in the paraphrase. The relative direction between the two objects is read from the sign of their bounding-box-centre difference along the relevant axis. The bonus rewards candidates whose relative direction agrees with the parsed $d_q$. Although bounding-box directions live in scene-specific world frames, in practice this signal is sufficient on 3D-FRONT, where most rooms align to consistent global frame conventions.

\paragraph{Combined structural bonus.}
Regardless of branch, the structural bonus $B(q, C) \in \mathbb{R}_{\geq 0}$ enters the final score with weight $\beta_{B}{=}0.3$, subtracted from the backbone cost. The router uses the LLM-predicted $\mathrm{axis}$ at no additional inference cost.

\subsection{LLM Verifier}
\label{sec:method:verifier}

\paragraph{Motivation.}
After the backbone and structural prior, the top-$3$ candidates are often nearly tied on subtle attributes that the $\{0, 0.5, 1\}$ indicators cannot resolve. A frozen LLM equipped with each candidate's full object list can disambiguate them holistically. Pilot studies showed that binary PASS/FAIL prompts collapse mild attribute drift and gross violation into a single FAIL, removing many correct top-$1$ candidates from the ranking. A continuous LLM confidence $\sigma(q, C) \in [0, 1]$, where $1$ means the candidate fully satisfies the edit and $0$ a clear violation, preserves the magnitude and lets a mildly imperfect but correct candidate outrank a clearly wrong one.

\paragraph{Design.}
We invoke a frozen LLM on each of the top-$3$ candidates ranked by the backbone and structural prior. The prompt asks it to output a continuous confidence $\sigma(q, C) \in [0, 1]$ together with a one-sentence rationale; the rationale is discarded. The verifier refines only positions one to three of the ranking; the tail is left untouched. This adds three LLM calls per query and roughly two seconds of wall-clock time.

\subsection{Final Score and Pipeline}
\label{sec:method:pipeline}

The final reranking score combines the three components additively:
\begin{equation}
\hat{s}(C) = \mathrm{OT}(q, C) - \beta_{B}\,B(q, C) - \beta_{V}\,\sigma(q, C)\cdot \mathbf{1}[C \in \text{top-}3],
\label{eq:final}
\end{equation}
with $\beta_{B}{=}0.3$ and $\beta_{V}{=}0.5$. The indicator $\mathbf{1}[C \in \text{top-}3]$ restricts the verifier to the three candidates with the lowest backbone-plus-prior score $\mathrm{OT}(q, C) - \beta_{B}\,B(q, C)$; the tail of $\mathcal{P}$ is left untouched.


\paragraph{Complexity.}
For one query, the cost-matrix and unbalanced-Sinkhorn solve are $\mathcal{O}(KG)$ with no matrix multiplication; the structural bonus is $\mathcal{O}(KG)$ for both the size-class and direction branches; the verifier issues exactly three LLM calls. End-to-end, the only parameters in the pipeline are the five scalars $(\varepsilon, \tau, \beta_{B}, \beta_{V}, K)$, all selected on the validation split before any test-set use.

\section{The 3D-CER Benchmark}
\label{sec:bench}

We release \textbf{3D-CER}, a benchmark of edit-conditioned queries over a $23{,}381$-room 3D-FRONT/3D-FUTURE corpus. To our knowledge, 3D-CER is the first 3D benchmark that combines edit-conditioned queries, multi-positive ground truth, CIRR-style hard subsets, and zero-target adversarials in a single resource (Table~\ref{tab:bench-comparison}).

\paragraph{Task formalism.}
A 3D-CER instance is a triplet $(R, t, \mathcal{T})$ where $R$ is a reference 3D room with object categories, styles, materials, and bounding boxes, $t$ is a natural-language edit, and $\mathcal{T} \subset \mathcal{C}$ is the set of multi-positive target rooms drawn from a fixed $23{,}381$-room corpus $\mathcal{C}$. In the hard-subset setting, methods rank $\mathcal{H} \cup \mathcal{T}$, where $\mathcal{H}$ is a $20$-room CIRR-style distractor pool mined per query from the same room type as the source~\citep{liu2021image}. About one fifth of test queries are zero-target ($\mathcal{T} = \emptyset$) and require abstention; these inject corpus-impossible attribute combinations such as a Scandinavian-style marble bathtub in a corpus without bathtubs.

\paragraph{Construction.}
3D-CER is built end-to-end with no human annotation. We programmatically mine candidate edit operators across five axes (style, material, spatial, geometric, compound) on $16{,}131$ valid source rooms, prompt GPT-5.1 for three short paraphrases per query in CIRR/CIRCO style, and pass each query through a five-layer programmatic validation that combines reverse-parse axis prediction, three-paraphrase axis consistency, source/target sanity checks, and JID/UUID leak detection ($98.5\%$ pass rate). Multi-positive ground truth is mined by running the operator predicate over the same-room-type slice of the corpus, and queries with no real positives become zero-target adversarials. Quality is independently audited by a held-out GPT-5.1 evaluator ($92.7\%$ distractor NO-rate on $500$ probes) and a cross-family check against Gemini-3-flash-preview-nothinking ($88.0\%$ verdict agreement on $1{,}478$ pairs). Full details of the seven construction phases and the audit protocols are deferred to Appendix~\ref{app:construction}.

\paragraph{Statistics and metric.}
3D-CER comprises $2{,}868$ train, $473$ val, and $1{,}622$ test queries over $23{,}381$ corpus rooms. The test split breaks down as $207$ STY, $152$ MAT, $388$ SPA, $475$ GEO, and $52$ CMP queries with ground truth, plus $348$ zero-target queries. The average multi-positive count is $14.7$ per query (median $5$). Following CIRCO~\citep{baldrati2023zero}, we adopt mAP@$10$ as the primary metric so that any residual axis-lenient false negatives do not unduly penalise a method that surfaces a valid but unlabelled positive.

\begin{table}[t]
\centering
\caption{Comparison with related benchmarks. \textbf{Edit}: pairs a reference and a modification text to retrieve a target. \textbf{Multi}: $\geq 2$ valid GTs per query. \textbf{Hard}: per-query distractor pool (CIRR-style). \textbf{Zero}: adversarial queries with no valid GT, requiring abstention. 3D-CER is the only benchmark that combines all four.}
\label{tab:bench-comparison}
\small
\begin{tabular}{lllrrcccc}
\toprule
Benchmark & Modality & Corpus & \#queries & & Edit & Multi & Hard & Zero \\
\midrule
ScanRefer~\citep{chen2020scanrefer}      & 3D scene  & 800        & 51{,}583   & & \xmark & \xmark & \xmark & \xmark \\
Multi3DRefer~\citep{zhang2023multi3drefer} & 3D scene & 800       & 61{,}926   & & \xmark & \cmark & \xmark & \cmark \\
CIRR~\citep{liu2021image}                  & 2D image  & 21{,}552  & 36{,}554   & & \cmark & \xmark & \cmark & \xmark \\
CIRCO~\citep{baldrati2023zero}           & 2D image  & 120{,}000 & 1{,}020    & & \cmark & \cmark & \xmark & \xmark \\
ChangeIt3D~\citep{achlioptas2023shapetalk} & 3D shape & ShapeNet & ${\sim}22$K & & \cmark & \xmark & \xmark & \xmark \\
\midrule
\textbf{3D-CER (ours)} & \textbf{3D scene} & \textbf{23{,}381} & \textbf{4{,}963} & & \cmark & \cmark & \cmark & \cmark \\
\bottomrule
\end{tabular}
\end{table}

\section{Experiments}
\label{sec:exp}

\subsection{Setup and Evaluation Protocol}
\label{sec:exp:setup}

We evaluate on the $1{,}622$-query test split of 3D-CER, comprising $1{,}274$ with-GT queries and $348$ zero-target queries. All experiments run on a single NVIDIA RTX $5090$ with random seed $42$. The base text encoder is MPNet (\texttt{all-mpnet-base-v2}). The frozen LLM used for entity parsing, paraphrase rewriting in the CIReVL baseline, and the verifier is GPT-5.1-2025-11-13; the cross-model audit uses Gemini-3-flash-preview-nothinking. Total combined wall-clock is approximately $4$ GPU-hours and the LLM API cost is approximately \$$30$. Hyperparameters $\tau{=}1.0$, $\beta_{\mathrm{B}}{=}0.3$, and $\beta_{V}{=}0.5$ are selected on the validation split before any test-set use. The hard-subset size is $20$.

We report four metrics. Recall@$1$, Recall@$5$, and Recall@$10$ measure recall over the $20$-distractor hard subset $\mathcal{H} \cup \mathcal{T}$, mirroring CIRR. mAP@$10$ measures mean Average Precision at $10$ over the same pool, accounting for multiple positives in CIRCO style. Abstention F1 is computed for zero-target queries: methods that emit no positive (or scores below a calibrated threshold) are correct, and F1 balances precision and recall of abstention. Per-axis breakdowns report all metrics across the five axes (STY, MAT, SPA, GEO, CMP) to expose method-axis interactions. mAP@$10$ is our primary metric due to the operator-level nature of multi-positive ground truth; bootstrap confidence intervals are reported in Appendix~\ref{app:bootstrap} where appropriate.

\subsection{Main Results}
\label{sec:exp:main}

Table~\ref{tab:main-results} reports R@$1$, R@$10$, and mAP@$10$ on the $1{,}274$ with-GT test queries against three training-free baselines that span three retrieval paradigms: dense embedding (caption-MPNet), LLM-rewrite (CIReVL~\citep{karthik2023vision}), and additive composition (Pic2Word~\citep{saito2023pic2word}). We additionally report intermediate ablations of CR-Refiner with one or two of its three components removed, so the source of the gain is visible at every step.

\begin{table}[t]
\centering
\caption{Main results on 3D-CER test, hard-subset reranking ($n{=}1{,}274$ with-GT). Top block: training-free baselines spanning dense embedding (Caption-MPNet), LLM rewrite (CIReVL), and additive composition (Pic2Word). Bottom block: cumulative components of CR-Refiner. \textit{Solver}: OT solver variant (\textit{add.}~$=$~additive cost, \textit{bal.}~$=$~balanced Sinkhorn, \textit{unb.}~$=$~unbalanced Sinkhorn). \textit{Prior}: structural prior with router strategy (\textit{bbox} $=$ bbox-only on every query, \textit{axis} $=$ axis-conditional). \textit{Ver.}: LLM verifier on the top three. \textbf{Bold} marks the full method.}
\label{tab:main-results}
\small
\begin{tabular}{lcccccc}
\toprule
Method & Solver & Prior & Ver. & R@1 & R@10 & mAP@10 \\
\midrule
Caption (MPNet)                  & ---  & ---  & ---  & 0.192 & 0.851 & 0.250 \\
CIReVL~\citep{karthik2023vision}   & ---  & ---  & ---  & 0.267 & 0.870 & 0.297 \\
Pic2Word~\citep{saito2023pic2word} & ---  & ---  & ---  & 0.272 & 0.863 & 0.316 \\
\midrule
CR-Refiner ablation             & add. & ---  & \xmark & 0.229 & 0.609 & 0.295 \\
                                & bal. & ---  & \xmark & 0.216 & 0.607 & 0.279 \\
                                & unb. & ---  & \xmark & 0.235 & 0.624 & 0.302 \\
                                & unb. & bbox & \xmark & 0.378 & 0.881 & 0.510 \\
                                & unb. & axis & \xmark & 0.585 & 0.947 & 0.651 \\
\textbf{CR-Refiner (full)}      & unb. & axis & \cmark & $\mathbf{0.639}$ & $\mathbf{0.951}$ & $\mathbf{0.663}$ \\
\bottomrule
\end{tabular}
\end{table}

Full CR-Refiner reaches R@$1{=}\mathbf{0.639}$ and mAP@$10{=}\mathbf{0.663}$, an absolute gain of $+36.7$ R@$1$ points and $+34.7$ mAP@$10$ points over the strongest baseline (Pic2Word). R@$10$ saturates near $0.951$ for the full method, so R@$1$ and mAP@$10$ are the more discriminating measures.

\subsection{Per-Axis Breakdown}
\label{sec:exp:peraxis}

Table~\ref{tab:peraxis} disaggregates R@$1$ by edit axis. We additionally include a non-LLM \textit{WL-only} baseline that scores candidates by a Weisfeiler--Lehman scene-graph kernel applied uniformly to all queries. This baseline isolates how much of the signal can be recovered from structured-metadata matching alone, without the OT backbone or the LLM verifier.

\begin{table}[t]
\centering
\caption{Per-axis R@$1$ on the full $1{,}274$ with-GT test pool (STY $207$ / MAT $152$ / SPA $388$ / GEO $475$ / CMP $52$). The \textit{WL-only} baseline scores candidates by a non-LLM Weisfeiler--Lehman scene-graph kernel applied uniformly.}
\label{tab:peraxis}
\begin{tabular}{lcccccc}
\toprule
Method & Overall & STY & MAT & SPA & GEO & CMP \\
\midrule
Caption (MPNet)                              & 0.192 & 0.357 & 0.171 & 0.204 & 0.120 & 0.173 \\
Pic2Word~\citep{saito2023pic2word}             & 0.272 & 0.275 & 0.224 & 0.526 & 0.078 & 0.288 \\
WL-only                                       & 0.351 & 0.367 & 0.191 & 0.562 & 0.253 & 0.077 \\
\textbf{Full CR-Refiner}                     & \textbf{0.639} & \textbf{0.725} & \textbf{0.513} & \textbf{0.546} & \textbf{0.724} & \textbf{0.577} \\
$\Delta$ vs.\ caption                         & $+44.7$ & $+36.8$ & $+34.2$ & $+34.2$ & $+60.4$ & $+40.4$ \\
\bottomrule
\end{tabular}
\end{table}

CR-Refiner improves R@$1$ over the caption baseline on every axis by between $34$ and $60$ absolute points, and matches or exceeds every text-conditioned baseline on every axis. On attribute axes (STY, MAT, GEO, CMP), the gain is driven by the OT backbone together with the bounding-box branch of the structural prior. On the spatial axis, the gain is delivered by the direction-bonus branch of the structural prior, reaching R@$1{=}0.546$, comparable to the non-LLM WL-only baseline at $0.562$. The WL-only baseline also confirms that 3D-CER is not trivially solvable by structured-metadata matching: WL-only beats caption overall ($0.351$ vs.\ $0.192$) but trails CR-Refiner by a wide margin ($0.351$ vs.\ $0.639$).

\subsection{Plug-in Compatibility}
\label{sec:exp:plugin}

Table~\ref{tab:plugin} reports the gain of CR-Refiner when stacked on three qualitatively distinct base retrievers under fixed-pool hard-subset reranking. The evaluation pool is the same released static hard subset used in Table~\ref{tab:main-results}; the difference is that this experiment uses a lightweight CR-Refiner configuration without the LLM verifier (\textit{OT backbone $+$ structural prior}, denoted $+$Ours) so that the plug-in cost remains constant across base retrievers. 

\begin{table}[t]
\centering
\small
\caption{Plug-in compatibility on three base retrievers. Same hard subset as Table~\ref{tab:main-results} ($n{=}1{,}274$ with-GT). $+$Ours is the verifier-free CR-Refiner; the verifier is excluded to isolate the retriever-agnostic component. Adding the verifier gives a further $+{\sim}5$ R@$1$ (see Table~\ref{tab:ablation}).}
\label{tab:plugin}
\begin{tabular}{lcccccc}
\toprule
& \multicolumn{3}{c}{R@$1$} & \multicolumn{3}{c}{mAP@$10$} \\
\cmidrule(lr){2-4}\cmidrule(lr){5-7}
Base retriever & alone & $+$Ours & $\Delta$ & alone & $+$Ours & $\Delta$ \\
\midrule
Caption (MPNet)                                & 0.185 & \textbf{0.578} & $+39.3$ & 0.248 & \textbf{0.609} & $+36.1$ \\
CIReVL~\citep{karthik2023vision}                 & 0.265 & \textbf{0.597} & $+33.2$ & 0.294 & \textbf{0.616} & $+32.2$ \\
Pic2Word~\citep{saito2023pic2word}              & 0.279 & \textbf{0.593} & $+31.4$ & 0.318 & \textbf{0.621} & $+30.3$ \\
\midrule
\multicolumn{1}{l}{\textit{Average $\Delta$ ($95\%$ CI)}} & & & $\mathbf{+34.6 \pm 4.0}$ & & & $\mathbf{+32.9 \pm 2.9}$ \\
\bottomrule
\end{tabular}
\end{table}

CR-Refiner uniformly delivers $+31$ to $+39$ R@$1$ points and $+30$ to $+36$ mAP@$10$ points across all three paradigms, indicating that the structural prior is retriever-agnostic under hard-subset reranking. We do not claim this property generalises to full-corpus retrieval, where recall rather than reranking becomes the bottleneck (Section~\ref{sec:exp:recallceiling}).

\subsection{Component Ablation}
\label{sec:exp:ablation}

Table~\ref{tab:ablation} reports two ablation studies on the full $1{,}274$ with-GT test queries. The top block removes one of the three components in turn while the other two are retained. The bottom block compares router variants of the structural prior. Each component's largest contribution coincides with its design axis. Removing the unbalanced solver (reverting to balanced Sinkhorn) costs $-10.5$ points on the material axis, where one-to-many asymmetry is most extreme: a query about a nightstand's material has nothing to say about the bed, the wardrobe, or the lamp, and a balanced solver dilutes the cost across all of them. Removing the structural prior is the largest single drop ($-34.2$ R@$1$ overall) and concentrates on the geometric axis ($-61.2$) and the spatial axis ($-35.8$), where bounding-box geometry and scene-graph adjacency carry the discriminative signal that the symbolic OT backbone cannot see. Removing the LLM verifier costs $-20.3$ points on the style axis, where holistic linguistic judgment correctly demotes candidates that match keyword-level style but fail when read in context.

\begin{table}[t]
\centering
\small
\caption{Ablation studies (R@$1$, $n{=}1{,}274$). \textbf{Top}: drop-one component, with the other two retained. \textbf{Bottom}: router variants. \textit{Bbox-only} forces the bbox branch on every query; \textit{oracle} uses ground-truth axis labels.}
\label{tab:ablation}
\begin{tabular}{lccccccc}
\toprule
Variant & Overall & STY & MAT & SPA & GEO & CMP & mAP@10 \\
\midrule
\multicolumn{8}{l}{\textit{Drop-one component ablation}} \\
\textbf{Full CR-Refiner}     & \textbf{0.639} & 0.725 & 0.513 & 0.546 & 0.724 & 0.577 & \textbf{0.663} \\
$-$ Unbalanced solver        & 0.613 & 0.729 & 0.408 & 0.552 & 0.691 & 0.500 & 0.633 \\
$-$ Structural prior         & 0.297 & 0.729 & 0.513 & 0.188 & 0.112 & 0.442 & 0.315 \\
$-$ LLM verifier             & 0.585 & 0.522 & 0.507 & 0.564 & 0.678 & 0.365 & 0.651 \\
\midrule
\multicolumn{8}{l}{\textit{Router variants}} \\
Bbox-only                    & 0.537 & 0.754 & 0.520 & 0.175 & 0.737 & 0.596 & --- \\
Oracle (upper bound)         & 0.640 & 0.754 & 0.520 & 0.546 & 0.737 & 0.596 & --- \\
\bottomrule
\end{tabular}
\end{table}

\begin{figure}[t]
  \centering
  \includegraphics[width=\linewidth]{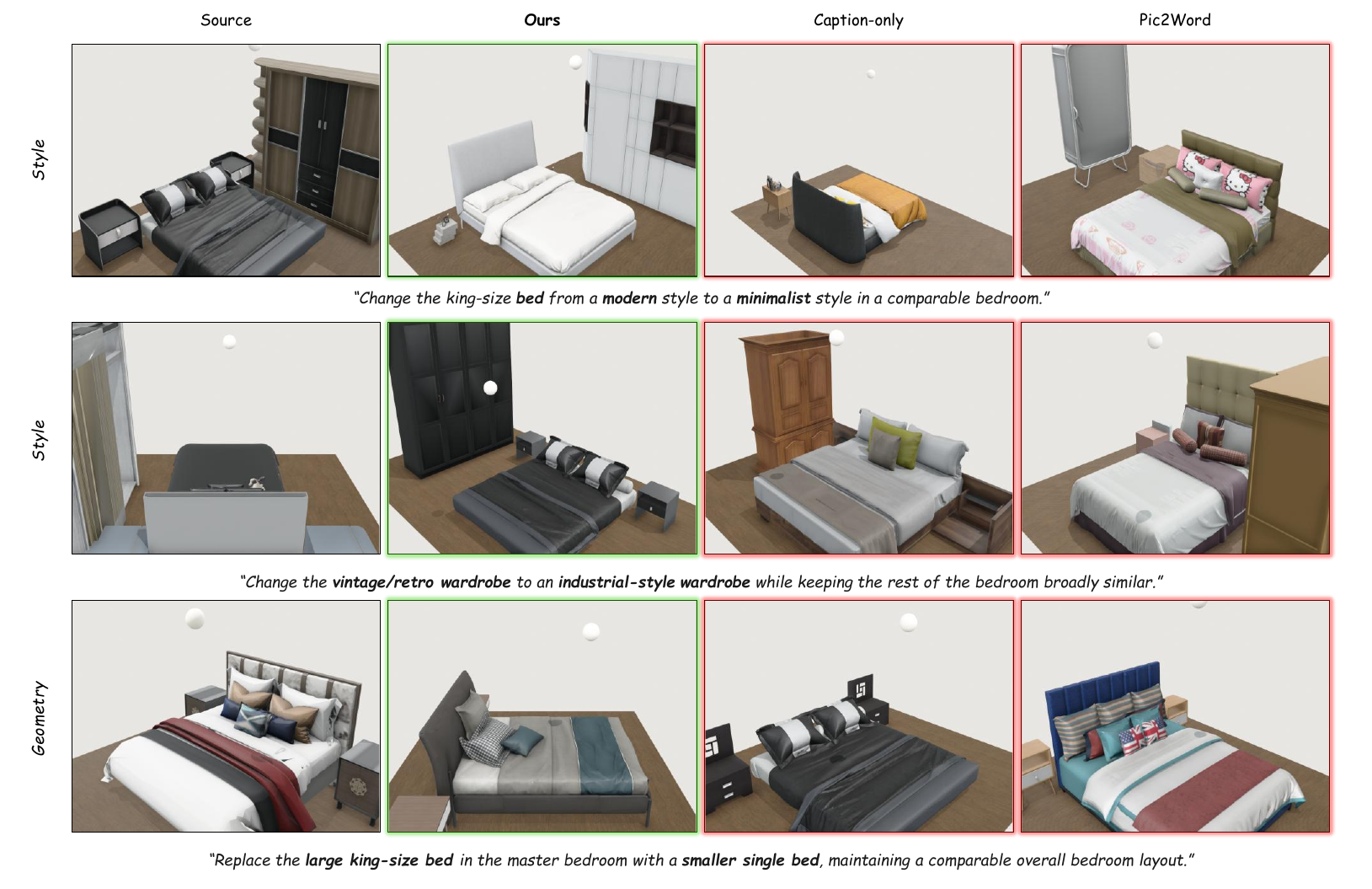}
  \vspace{-13pt}
  \caption{\textbf{Qualitative comparison on 3D-CER hard subsets.} Source room (column~$1$) and top-$1$ retrievals from CR-Refiner (column~$2$, green border), Caption-MPNet (column~$3$, red border), and Pic2Word (column~$4$, red border). The edit axis is labelled at the left of each row; the edit instruction appears in italics below.}
  \label{fig:qualitative}
  \vspace{-10pt}
\end{figure}



\paragraph{Routing.}
Axis-conditional routing lifts overall R@$1$ by $+10.2$ points over the bbox-only variant, almost entirely from the spatial axis ($+37.1$ points), where the bbox-only variant disables the direction bonus that SPA queries depend on. The oracle-routed upper bound at $0.640$ is essentially identical to the parser-routed full method at $0.639$, so a learned router is not necessary.

\subsection{Full-Corpus Recall Ceiling}
\label{sec:exp:recallceiling}

All numbers above are fixed-pool reranking. Since a reranker only re-orders the base retriever's output, we measure the full-corpus recall ceiling on $200$ stratified queries. Even the strongest base retriever (CIReVL) recalls only $88.0\%$ of positives at $K{=}1000$ and $7.5\%$ at $K{=}1$. Feeding CR-Refiner a top-$K$ from this corpus produces a modest lift at small $K$ that disappears at large $K$ as the structural prior dilutes against an increasingly noisy pool. CR-Refiner is therefore effective for candidate disambiguation but not sufficient on broad pools, and closing the full-corpus gap is a candidate-generation problem we leave to future work.

\vspace{-10pt}
\subsection{Qualitative Analysis}
\label{sec:exp:qualitative}
\vspace{-5pt}
Figure~\ref{fig:qualitative} shows top-$1$ retrievals on three hard-subset queries, two STY edits and one GEO edit. In every row, only CR-Refiner returns a room where the slot named in the edit text changes as instructed (bed style in Row~1, wardrobe style in Row~2, and bed size class in Row~3), while the remaining furniture stays close to the source. Caption-MPNet matches surface keywords but selects rooms with visibly different layouts, and Pic2Word retrieves rooms where the edited slot is occluded or missing from view, consistent with additive composition failing to localise the edit on a named object.
\vspace{-10pt}

\section{Limitations}
\label{sec:limit}
\vspace{-5pt}
All headline numbers are reranking over a $20$-distractor hard subset rather than full-corpus retrieval, and the recall ceiling in Section~\ref{sec:exp:recallceiling} bounds any downstream reranker by the candidate-generation stage. Both paraphrases and the multi-positive predicate operate at the operator level following the CIRR/CIRCO convention, which we mitigate by adopting mAP@$10$ as the primary metric. The LLM verifier adds roughly two seconds per query, which is acceptable offline but limits real-time deployment; the OT backbone and structural prior alone run on CPU at over $500$ queries per second. Each of these scopes points to a complementary research direction left for future work.

\vspace{-10pt}
\section{Conclusion}
\label{sec:conclusion}
\vspace{-5pt}
We presented CR-Refiner, a training-free reranker for edit-conditioned 3D scene retrieval that frames the problem as an unbalanced optimal-transport problem over an asymmetric cost matrix, complemented by an axis-conditional structural prior and a continuous-confidence LLM verifier. We additionally release 3D-CER, a benchmark of edit-conditioned queries over a $23$K-room indoor corpus with multi-positive ground truth, hard subsets, and zero-target adversarials. CR-Refiner consistently improves hard-subset retrieval over training-free baselines on every edit axis, and the released benchmark and code support reproducible follow-up work on 3D-aware retrieval.

\bibliographystyle{unsrtnat}
\bibliography{references}

@inproceedings{liu2021image,
  title={Image retrieval on real-life images with pre-trained vision-and-language models},
  author={Liu, Zheyuan and Rodriguez-Opazo, Cristian and Teney, Damien and Gould, Stephen},
  booktitle={Proceedings of the IEEE/CVF international conference on computer vision},
  pages={2125--2134},
  year={2021}
}

@inproceedings{baldrati2023zero,
  title={Zero-shot composed image retrieval with textual inversion},
  author={Baldrati, Alberto and Agnolucci, Lorenzo and Bertini, Marco and Del Bimbo, Alberto},
  booktitle={Proceedings of the IEEE/CVF international conference on computer vision},
  pages={15338--15347},
  year={2023}
}

@article{karthik2023vision,
  title={Vision-by-language for training-free compositional image retrieval},
  author={Karthik, Shyamgopal and Roth, Karsten and Mancini, Massimiliano and Akata, Zeynep},
  journal={arXiv preprint arXiv:2310.09291},
  year={2023}
}

@inproceedings{saito2023pic2word,
  title={Pic2word: Mapping pictures to words for zero-shot composed image retrieval},
  author={Saito, Kuniaki and Sohn, Kihyuk and Zhang, Xiang and Li, Chun-Liang and Lee, Chen-Yu and Saenko, Kate and Pfister, Tomas},
  booktitle={Proceedings of the IEEE/CVF conference on computer vision and pattern recognition},
  pages={19305--19314},
  year={2023}
}

@article{liu2023openshape,
  title={Openshape: Scaling up 3d shape representation towards open-world understanding},
  author={Liu, Minghua and Shi, Ruoxi and Kuang, Kaiming and Zhu, Yinhao and Li, Xuanlin and Han, Shizhong and Cai, Hong and Porikli, Fatih and Su, Hao},
  journal={Advances in neural information processing systems},
  volume={36},
  pages={44860--44879},
  year={2023}
}

@article{zhou2023uni3d,
  title={Uni3d: Exploring unified 3d representation at scale},
  author={Zhou, Junsheng and Wang, Jinsheng and Ma, Baorui and Liu, Yu-Shen and Huang, Tiejun and Wang, Xinlong},
  journal={arXiv preprint arXiv:2310.06773},
  year={2023}
}

@inproceedings{xue2024ulip,
  title={Ulip-2: Towards scalable multimodal pre-training for 3d understanding},
  author={Xue, Le and Yu, Ning and Zhang, Shu and Panagopoulou, Artemis and Li, Junnan and Mart{\'\i}n-Mart{\'\i}n, Roberto and Wu, Jiajun and Xiong, Caiming and Xu, Ran and Niebles, Juan Carlos and others},
  booktitle={Proceedings of the IEEE/CVF Conference on Computer Vision and Pattern Recognition},
  pages={27091--27101},
  year={2024}
}

@inproceedings{chen2020scanrefer,
  title={Scanrefer: 3d object localization in rgb-d scans using natural language},
  author={Chen, Dave Zhenyu and Chang, Angel X and Nie{\ss}ner, Matthias},
  booktitle={European conference on computer vision},
  pages={202--221},
  year={2020},
  organization={Springer}
}

@inproceedings{zhang2023multi3drefer,
  title={Multi3drefer: Grounding text description to multiple 3d objects},
  author={Zhang, Yiming and Gong, ZeMing and Chang, Angel X},
  booktitle={Proceedings of the IEEE/CVF International Conference on Computer Vision},
  pages={15225--15236},
  year={2023}
}

@inproceedings{jia2024sceneverse,
  title={Sceneverse: Scaling 3d vision-language learning for grounded scene understanding},
  author={Jia, Baoxiong and Chen, Yixin and Yu, Huangyue and Wang, Yan and Niu, Xuesong and Liu, Tengyu and Li, Qing and Huang, Siyuan},
  booktitle={European Conference on Computer Vision},
  pages={289--310},
  year={2024},
  organization={Springer}
}

@inproceedings{achlioptas2023shapetalk,
  title={ShapeTalk: A language dataset and framework for 3d shape edits and deformations},
  author={Achlioptas, Panos and Huang, Ian and Sung, Minhyuk and Tulyakov, Sergey and Guibas, Leonidas},
  booktitle={Proceedings of the IEEE/CVF conference on computer vision and pattern recognition},
  pages={12685--12694},
  year={2023}
}

@article{chizat2018scaling,
  title={Scaling algorithms for unbalanced optimal transport problems},
  author={Chizat, Lenaic and Peyr{\'e}, Gabriel and Schmitzer, Bernhard and Vialard, Fran{\c{c}}ois-Xavier},
  journal={Mathematics of computation},
  volume={87},
  number={314},
  pages={2563--2609},
  year={2018}
}

@misc{zheng2025editroomllmparameterizedgraphdiffusion,
      title={EditRoom: LLM-parameterized Graph Diffusion for Composable 3D Room Layout Editing}, 
      author={Kaizhi Zheng and Xiaotong Chen and Xuehai He and Jing Gu and Linjie Li and Zhengyuan Yang and Kevin Lin and Jianfeng Wang and Lijuan Wang and Xin Eric Wang},
      year={2025},
      eprint={2410.12836},
      archivePrefix={arXiv},
      primaryClass={cs.GR},
      url={https://arxiv.org/abs/2410.12836}, 
}

@misc{azuma2022scanqa3dquestionanswering,
      title={ScanQA: 3D Question Answering for Spatial Scene Understanding}, 
      author={Daichi Azuma and Taiki Miyanishi and Shuhei Kurita and Motoaki Kawanabe},
      year={2022},
      eprint={2112.10482},
      archivePrefix={arXiv},
      primaryClass={cs.CV},
      url={https://arxiv.org/abs/2112.10482}, 
}

@InProceedings{10.1007/978-3-030-58452-8_25,
author="Achlioptas, Panos
and Abdelreheem, Ahmed
and Xia, Fei
and Elhoseiny, Mohamed
and Guibas, Leonidas",
editor="Vedaldi, Andrea
and Bischof, Horst
and Brox, Thomas
and Frahm, Jan-Michael",
title="ReferIt3D: Neural Listeners for Fine-Grained 3D Object Identification in Real-World Scenes",
booktitle="Computer Vision -- ECCV 2020",
year="2020",
publisher="Springer International Publishing",
address="Cham",
pages="422--440",
isbn="978-3-030-58452-8"
}

@misc{vo2018composingtextimageimage,
      title={Composing Text and Image for Image Retrieval - An Empirical Odyssey}, 
      author={Nam Vo and Lu Jiang and Chen Sun and Kevin Murphy and Li-Jia Li and Li Fei-Fei and James Hays},
      year={2018},
      eprint={1812.07119},
      archivePrefix={arXiv},
      primaryClass={cs.CV},
      url={https://arxiv.org/abs/1812.07119}, 
}

@inproceedings{baldrati2022effective,
  title={Effective conditioned and composed image retrieval combining clip-based features},
  author={Baldrati, Alberto and Bertini, Marco and Uricchio, Tiberio and Del Bimbo, Alberto},
  booktitle={Proceedings of the IEEE/CVF conference on computer vision and pattern recognition},
  pages={21466--21474},
  year={2022}
}

@misc{gu2024languageonlyefficienttrainingzeroshot,
      title={Language-only Efficient Training of Zero-shot Composed Image Retrieval}, 
      author={Geonmo Gu and Sanghyuk Chun and Wonjae Kim and Yoohoon Kang and Sangdoo Yun},
      year={2024},
      eprint={2312.01998},
      archivePrefix={arXiv},
      primaryClass={cs.CV},
      url={https://arxiv.org/abs/2312.01998}, 
}

@misc{
gu2024compodiff,
title={CompoDiff: Versatile Composed Image Retrieval With Latent Diffusion},
author={Geonmo Gu and Sanghyuk Chun and Wonjae Kim and HeeJae Jun and Yoohoon Kang and Sangdoo Yun},
year={2024},
url={https://openreview.net/forum?id=0NruoU6s5Z}
}

@misc{zhu20233dvistapretrainedtransformer3d,
      title={3D-VisTA: Pre-trained Transformer for 3D Vision and Text Alignment}, 
      author={Ziyu Zhu and Xiaojian Ma and Yixin Chen and Zhidong Deng and Siyuan Huang and Qing Li},
      year={2023},
      eprint={2308.04352},
      archivePrefix={arXiv},
      primaryClass={cs.CV},
      url={https://arxiv.org/abs/2308.04352}, 
}

@misc{hong20233dllminjecting3dworld,
      title={3D-LLM: Injecting the 3D World into Large Language Models}, 
      author={Yining Hong and Haoyu Zhen and Peihao Chen and Shuhong Zheng and Yilun Du and Zhenfang Chen and Chuang Gan},
      year={2023},
      eprint={2307.12981},
      archivePrefix={arXiv},
      primaryClass={cs.CV},
      url={https://arxiv.org/abs/2307.12981}, 
}

@misc{xu2024pointllmempoweringlargelanguage,
      title={PointLLM: Empowering Large Language Models to Understand Point Clouds}, 
      author={Runsen Xu and Xiaolong Wang and Tai Wang and Yilun Chen and Jiangmiao Pang and Dahua Lin},
      year={2024},
      eprint={2308.16911},
      archivePrefix={arXiv},
      primaryClass={cs.CV},
      url={https://arxiv.org/abs/2308.16911}, 
}

@misc{miao2024scenegraphloccrossmodalcoarsevisual,
      title={SceneGraphLoc: Cross-Modal Coarse Visual Localization on 3D Scene Graphs}, 
      author={Yang Miao and Francis Engelmann and Olga Vysotska and Federico Tombari and Marc Pollefeys and Dániel Béla Baráth},
      year={2024},
      eprint={2404.00469},
      archivePrefix={arXiv},
      primaryClass={cs.CV},
      url={https://arxiv.org/abs/2404.00469}, 
}

@misc{zhang2021pointclippointcloudunderstanding,
      title={PointCLIP: Point Cloud Understanding by CLIP}, 
      author={Renrui Zhang and Ziyu Guo and Wei Zhang and Kunchang Li and Xupeng Miao and Bin Cui and Yu Qiao and Peng Gao and Hongsheng Li},
      year={2021},
      eprint={2112.02413},
      archivePrefix={arXiv},
      primaryClass={cs.CV},
      url={https://arxiv.org/abs/2112.02413}, 
}

@misc{tang2024diffuscenedenoisingdiffusionmodels,
      title={DiffuScene: Denoising Diffusion Models for Generative Indoor Scene Synthesis}, 
      author={Jiapeng Tang and Yinyu Nie and Lev Markhasin and Angela Dai and Justus Thies and Matthias Nießner},
      year={2024},
      eprint={2303.14207},
      archivePrefix={arXiv},
      primaryClass={cs.CV},
      url={https://arxiv.org/abs/2303.14207}, 
}

@misc{lin2024instructsceneinstructiondriven3dindoor,
      title={InstructScene: Instruction-Driven 3D Indoor Scene Synthesis with Semantic Graph Prior}, 
      author={Chenguo Lin and Yadong Mu},
      year={2024},
      eprint={2402.04717},
      archivePrefix={arXiv},
      primaryClass={cs.CV},
      url={https://arxiv.org/abs/2402.04717}, 
}


\appendix

\section{Pseudocode of CR-Refiner}
\label{app:algorithm}

Algorithm~\ref{alg:cr-refiner} summarises the full inference procedure of CR-Refiner. Each step references the corresponding component of the main text. Line 1 invokes the LLM parser (Section~\ref{sec:method:prelim}). Line 4 builds the cost matrix and solves the unbalanced Sinkhorn (Section~\ref{sec:method:backbone}). Line 5 computes the axis-conditional structural bonus (Section~\ref{sec:method:prior}). Line 7 runs the LLM verifier on the top three candidates only (Section~\ref{sec:method:verifier}).

\begin{algorithm}[h]
\caption{CR-Refiner reranking}
\label{alg:cr-refiner}
\begin{algorithmic}[1]
\Require reference room $R$, edit text $t$, base retriever $\mathcal{R}$, frozen LLM $\Phi$, corpus $\mathcal{C}$, top-$K$ size $K$
\State $q \gets \Phi(R, t)$ \Comment{parse to query entity}
\State $\mathcal{P} \gets \mathcal{R}(R, t, \mathcal{C}, K)$ \Comment{base retriever's top-$K$}
\For{$C \in \mathcal{P}$}
   \State Build cost matrix $\mathbf{M}$ (Eq.~\ref{eq:costmatrix}); solve Eq.~\ref{eq:unbalanced} $\to \mathrm{OT}(q, C)$
   \State Route by $q.\mathrm{axis}$; compute structural bonus $B(q, C)$
\EndFor
\State Compute partial scores $\mathrm{OT}(q, C) - \beta_{\mathrm{B}}\,B(q, C)$ for all $C \in \mathcal{P}$
\State $\sigma(q, C) \gets \Phi(q, C)$ for the top three candidates by partial score
\State Compute final $\hat{s}(C)$ via Eq.~\ref{eq:final} for all $C \in \mathcal{P}$
\State \Return $\mathcal{P}$ sorted by $\hat{s}$ in ascending order
\end{algorithmic}
\end{algorithm}

\paragraph{Hyperparameters.}
The full pipeline contains five scalar hyperparameters. The entropic regularisation weight $\varepsilon{=}0.01$ and the KL slack $\tau{=}1.0$ control the unbalanced Sinkhorn (Eq.~\ref{eq:unbalanced}). The structural-bonus weight $\beta_{B}{=}0.3$ scales the structural prior in the final score (Eq.~\ref{eq:final}). The verifier weight $\beta_{V}{=}0.5$ scales the LLM-verifier confidence (Eq.~\ref{eq:final}). The candidate pool size is $K = |\mathcal{H} \cup \mathcal{T}|$, where $|\mathcal{H}|=20$ caption-NN distractors are fixed per query (\S\ref{sec:bench}) and $|\mathcal{T}|$ is the per-query multi-positive count (median $5$, mean $14.7$). All five hyperparameters are selected on the validation split before any test-set use.

\section{3D-CER Construction Pipeline}
\label{app:construction}

3D-CER is built end-to-end with no human annotation. Correctness is enforced by programmatic predicates and an independent cross-model audit. The pipeline consists of seven phases, summarised below.

\paragraph{Phase 1: Edit-operator mining.}
For each of the $16{,}131$ valid source rooms (those containing three to twelve furniture objects, after filtering empty and over-cluttered scenes from the $23{,}381$-room corpus), we programmatically generate up to $30$ candidate edit operators across the five axes (style, material, spatial, geometric, compound). Each operator is a structured object specifying axis, target slot, source attribute, target attribute, and predicate. The mining yields over $460{,}000$ (room, operator) pairs.

\paragraph{Phase 2: Synthetic target rendering for LLM grounding.}
We balance-sample $2{,}000$ candidates per axis ($10{,}000$ total) and synthesise a target room via mesh swap for style and material edits, spatial shift for spatial edits, or category replacement for geometric edits. The synthetic target is used only as visual grounding for the GPT-5.1 paraphrase generator in Phase 3 and is \emph{not} part of the released corpus or ground truth. After compatibility filtering, $8{,}966$ candidates are retained.

\paragraph{Phase 3: Paraphrase generation.}
Given each (source, operator) pair, we prompt GPT-5.1 to produce three short paraphrases per query in three registers (formal, colloquial, affective), following the CIRR/CIRCO convention for composed-retrieval text. Each paraphrase is one or two sentences, names the key edited object, and describes the axis-specific change with at most a soft preservation hint such as ``in roughly the same layout'' or ``broadly similar''. Hard constraints in the prompt forbid object-ID leakage and coordinate disclosure. All $8{,}966$ candidates produce valid paraphrases. The exact prompt template is reproduced in Appendix~\ref{app:prompts}.

\paragraph{Phase 4: Five-layer programmatic validation.}
Each query passes five binary checks before becoming a release candidate.
\textbf{(L1)} Reverse-parse axis prediction. We parse the paraphrase back into a query entity and verify that the parsed axis matches the source operator axis.
\textbf{(L2)} Three-paraphrase axis consistency. All three paraphrases of the same query must agree on the axis label.
\textbf{(L3)} Source-as-positive sanity. The original source room must satisfy the edit predicate when the operator's source attribute is replaced by the target attribute.
\textbf{(L4)} Target-as-positive sanity. The synthetic target room from Phase 2 must pass the predicate.
\textbf{(L5)} JID/UUID leak detection. No paraphrase contains 3D-FUTURE asset IDs that could be used as shortcuts.
The end-to-end pass rate is $8{,}834 / 8{,}966 = 98.5\%$.

\paragraph{Phase 5: Multi-positive ground-truth mining.}
For each surviving query, we run the operator predicate over the same-room-type slice of the $23{,}381$-room corpus to mine multi-positive ground truth. Of the $8{,}834$ validated queries, $3{,}923$ ($44.4\%$) yield $\geq 1$ real positive in the corpus and form the with-GT pool. The remaining $4{,}911$ queries yield zero positives and become \emph{zero-target adversarials}, a free byproduct of strict predicate matching. The average multi-positive count is $14.7$ per query (median $5$). Concretely, a zero-target adversarial such as ``Switch the bathtub to Scandinavian-style marble'' produces no positive when the corpus contains no bathtub object, regardless of the requested style or material; methods must learn to recognise this gap and abstain rather than return a falsely confident match.

\paragraph{Phase 6: AI-judge quality control.}
A randomly sampled set of $500$ queries with $6$ probes each (3 positive candidates and 3 distractor candidates per query) is judged by GPT-5.1 in a held-out evaluator role. The distractor NO-rate is $92.7\%$, indicating that hard-subset distractors are reliably distinguishable from positives. We additionally run a cross-family independence check using Gemini-3-flash-preview-nothinking on $1{,}478$ (query, candidate) pairs. The two model families agree on $88.0\%$ of verdicts, with the confusion matrix dominated by easy distractor rejections (NO/NO: $1{,}203$ pairs). Cross-family agreement on positives is correspondingly modest, consistent with the operator-level multi-positive predicate (one axis-specific edit, applied to the named slot, in a 3D scene broadly comparable to the source).

\paragraph{Phase 7: Stratified split.}
We split queries into train ($2{,}868$), validation ($473$), and test ($1{,}622$) sets, stratified jointly by edit axis and source room type. Roughly $21\%$ of test queries are zero-target. The test split breaks down as $207$ STY, $152$ MAT, $388$ SPA, $475$ GEO, and $52$ CMP queries with ground truth, plus $348$ zero-target queries.

\paragraph{Hard subset construction.}
For each release-candidate query, we mine $|\mathcal{H}|=20$ caption-nearest-neighbour distractors from the same-room-type slice of the corpus. Distractors are ranked by cosine similarity between the source room's caption embedding (MPNet \texttt{all-mpnet-base-v2}) and each candidate room's caption embedding, with all multi-positive ground-truth rooms $\mathcal{T}$ excluded so that $\mathcal{H} \cap \mathcal{T} = \emptyset$. The hard subset is fixed per query and released as part of the static benchmark.

\paragraph{Example paraphrases.}
We give one paraphrase per axis to illustrate the dataset's textual style.
\begin{itemize}
\item \textbf{STY:} \textit{``Change the king-size bed in this second bedroom from a light-luxury style to a vintage/retro style, with the room's overall composition broadly comparable.''}
\item \textbf{MAT:} \textit{``Change the single bed's material from cloth to plywood in a comparable second bedroom configuration.''}
\item \textbf{SPA:} \textit{``Place the nightstand on the right of the bed instead of the left, keeping the rest of the room broadly similar.''}
\item \textbf{GEO:} \textit{``Replace the large king-size bed with a smaller single bed in a comparable second bedroom layout.''}
\item \textbf{CMP:} \textit{``Switch the wardrobe to a Japanese style and move it next to the window, otherwise keep the bedroom broadly similar.''}
\end{itemize}

\section{Bootstrap Confidence Intervals}
\label{app:bootstrap}

We compute $95\%$ bootstrap confidence intervals for the three headline metrics in Table~\ref{tab:main-results} and the plug-in deltas in Table~\ref{tab:plugin}. Bootstraps are computed over $1{,}000$ resamples of the $1{,}274$-query test split, with the same evaluation pool per query as in the main text. Table~\ref{tab:bootstrap-main} reports CIs for the full method against the strongest baseline, and Table~\ref{tab:bootstrap-plugin} reports CIs for the plug-in deltas across base retrievers. The CIs do not overlap between the full method and any text-conditioned baseline on R@$1$ or mAP@$10$, confirming that the gap is not within bootstrap noise.

\begin{table}[h]
\centering
\small
\caption{$95\%$ bootstrap CIs for headline metrics ($n{=}1{,}274$, $1{,}000$ resamples).}
\label{tab:bootstrap-main}
\begin{tabular}{lccc}
\toprule
Method & R@$1$ & R@$10$ & mAP@$10$ \\
\midrule
Caption (MPNet)            & $0.192$ [$0.171$, $0.213$] & $0.851$ [$0.832$, $0.870$] & $0.250$ [$0.231$, $0.270$] \\
Pic2Word~\citep{saito2023pic2word} & $0.272$ [$0.250$, $0.297$] & $0.863$ [$0.844$, $0.882$] & $0.316$ [$0.292$, $0.339$] \\
\textbf{Full CR-Refiner}   & $\mathbf{0.639}$ [$0.612$, $0.665$] & $\mathbf{0.951}$ [$0.939$, $0.962$] & $\mathbf{0.663}$ [$0.638$, $0.687$] \\
\bottomrule
\end{tabular}
\end{table}

\begin{table}[h]
\centering
\small
\caption{$95\%$ bootstrap CIs for plug-in $\Delta$ in Table~\ref{tab:plugin}.}
\label{tab:bootstrap-plugin}
\begin{tabular}{lcc}
\toprule
Base retriever & $\Delta$ R@$1$ & $\Delta$ mAP@$10$ \\
\midrule
Caption (MPNet)                  & $+39.3$ [$+35.7$, $+42.9$] & $+36.1$ [$+32.4$, $+39.8$] \\
CIReVL~\citep{karthik2023vision} & $+33.2$ [$+29.8$, $+36.6$] & $+32.2$ [$+28.7$, $+35.7$] \\
Pic2Word~\citep{saito2023pic2word} & $+31.4$ [$+28.0$, $+34.8$] & $+30.3$ [$+26.9$, $+33.7$] \\
\midrule
\textbf{Average}                 & $\mathbf{+34.6 \pm 4.0}$ & $\mathbf{+32.9 \pm 2.9}$ \\
\bottomrule
\end{tabular}
\end{table}

\section{Full-Corpus Recall Ceiling and Pipeline Decomposition}
\label{app:pipeline-decomp}

We provide two diagnostic tables that complement the brief discussion in Section~\ref{sec:exp:recallceiling}. Table~\ref{tab:recallceiling} reports full-corpus Recall@$K$ for each base retriever, and Table~\ref{tab:kdecomp} reports end-to-end pipeline R@$1$ when CR-Refiner reranks the base retriever's top-$K$ from the full corpus.

\begin{table}[h]
\centering
\small
\caption{Full-corpus Recall@$K$ on the $23{,}381$-room corpus ($n{=}200$).}
\label{tab:recallceiling}
\begin{tabular}{lccccc}
\toprule
Method & R@1 & R@10 & R@100 & R@500 & R@1000 \\
\midrule
Caption (MPNet)        & 0.000 & 0.055 & 0.200 & 0.520 & 0.665 \\
Pic2Word (additive)    & 0.045 & 0.190 & 0.465 & 0.740 & 0.835 \\
CIReVL (LLM rewrite)   & \textbf{0.075} & \textbf{0.275} & \textbf{0.560} & \textbf{0.805} & \textbf{0.880} \\
\bottomrule
\end{tabular}
\end{table}

\begin{figure}[t]
  \centering
  \includegraphics[width=\linewidth]{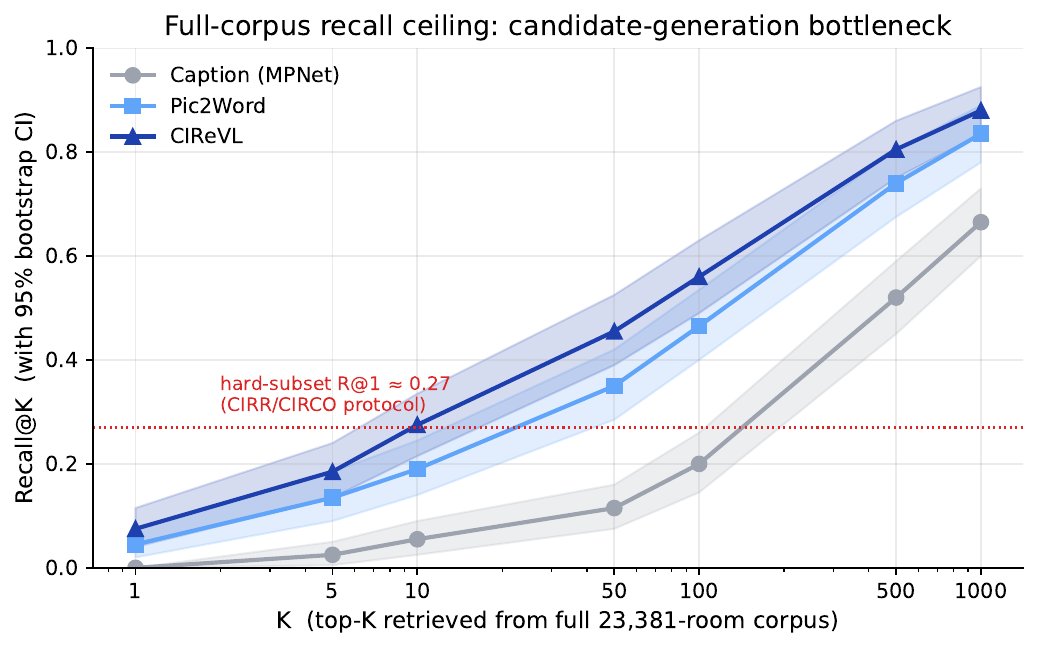}
  \caption{\textbf{Full-corpus recall ceiling on the $23{,}381$-room corpus} ($n{=}200$). Recall@$K$ for the three base retrievers across $K \in \{1, 5, 10, 50, 100, 500, 1000\}$, with $95\%$ bootstrap confidence intervals shaded. The dotted line marks the hard-subset R@$1$ regime ($\approx 0.27$) reported in Section~\ref{sec:exp:main}, illustrating how far full-corpus recall lags the curated-pool numbers at small $K$. Visualisation of Table~\ref{tab:recallceiling}.}
  \label{fig:recall-ceiling}
\end{figure}

The recall ceiling rises slowly with $K$ (Figure~\ref{fig:recall-ceiling}). Even the strongest base retriever recalls only $88.0\%$ of positives at $K{=}1000$ (the top $4.3\%$ of the corpus), and the dense caption retriever recalls $66.5\%$. The base-retriever R@$1$ on the full corpus is essentially zero for caption-MPNet and only $0.045$ to $0.075$ for the strongest text-conditioned retrievers, even though the same retrievers reach hard-subset R@$1$ of $0.19$ to $0.27$ in Section~\ref{sec:exp:main}. The two-orders-of-magnitude gap shows that hard-subset reranking and full-corpus retrieval are different regimes rather than two points on the same curve.

\begin{table}[h]
\centering
\small
\caption{End-to-end full-corpus pipeline R@$1$ by base-retriever pool size $K$ ($n{=}200$). The verifier is disabled here to avoid conflating the diagnostic with per-query inference cost at scale.}
\label{tab:kdecomp}
\begin{tabular}{lcccccc}
\toprule
Base retriever & Base R@1 & $K{=}10$ & $K{=}50$ & $K{=}100$ & $K{=}500$ & $K{=}1000$ \\
\midrule
Caption (MPNet) $+$ CR-Refiner    & 0.000 & 0.010 & 0.010 & 0.010 & 0.005 & 0.005 \\
Pic2Word $+$ CR-Refiner           & 0.045 & \textbf{0.055} & 0.025 & 0.040 & 0.015 & 0.015 \\
\bottomrule
\end{tabular}
\end{table}

\begin{figure}[t]
  \centering
  \includegraphics[width=\linewidth]{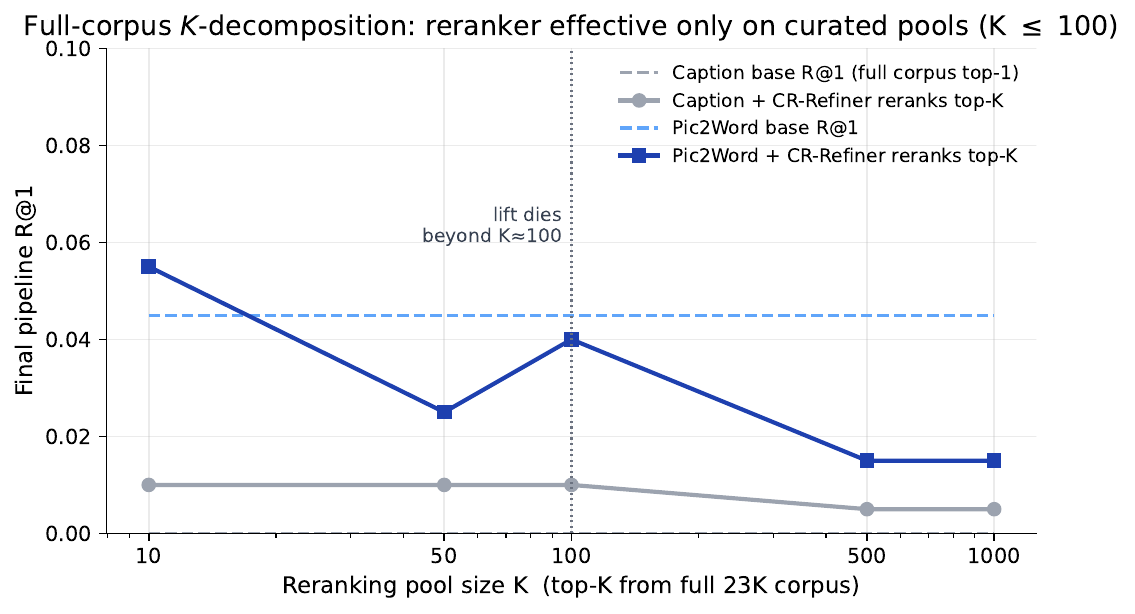}
  \caption{\textbf{Pipeline R@$1$ as a function of reranking pool size $K$} ($n{=}200$). Solid lines are the end-to-end pipeline R@$1$ when CR-Refiner reranks the base retriever's top-$K$ from the full $23$K-room corpus; dashed lines are the corresponding base retriever's full-corpus top-$1$ for reference. The lift over the base retriever survives at $K{\leq}100$ and dies beyond $K{\approx}100$ as the structural prior dilutes against an increasingly noisy pool. Visualisation of Table~\ref{tab:kdecomp}.}
  \label{fig:k-decomp}
\end{figure}

The shape of the curve is the substantive finding (Figure~\ref{fig:k-decomp}). At small $K$ from $10$ to $100$, reranking provides a modest lift over the base retriever's top-$1$, with overlapping bootstrap intervals that we read as suggestive rather than conclusive. At large $K$ such as $500$ or $1000$, the structural prior dilutes against an increasingly noisy pool and the lift disappears. We treat this as a scope diagnostic that makes the reranker's operating regime explicit. Closing the full-corpus recall gap is a candidate-generation problem requiring a stronger 3D-aware retriever, for example a frozen 3D foundation encoder such as Uni3D~\citep{zhou2023uni3d} or OpenShape~\citep{liu2023openshape} adapted to scene-level composition, which is complementary future work.

\section{LLM Prompts}
\label{app:prompts}

We reproduce the prompts used for entity parsing, paraphrase generation, and verification. All prompts are sent to GPT-5.1-2025-11-13 with temperature $0$ and a fixed system message asking for JSON-only output.

\paragraph{Entity parser (Section~\ref{sec:method:prelim}).}
\begin{quote}\small\itshape
You are a structured-edit parser for indoor 3D scenes. Given a reference room description and an edit instruction, output a JSON object with fields: \texttt{category} (target object), \texttt{style} (target style or null), \texttt{material} (target material or null), \texttt{spatial} (left/right/front/behind/near/far or null), and \texttt{axis} (one of STY, MAT, SPA, GEO, CMP). Do not output any extra text.

Reference room: \texttt{\{spec\}}\\
Edit instruction: \texttt{\{paraphrase\}}\\
JSON:
\end{quote}

\paragraph{Paraphrase generator (Phase 3).}
\begin{quote}\small\itshape
You are writing natural-language editing instructions for an interior design retrieval task. Given a source room and a structured edit operator, generate three short paraphrases of the edit in three registers: formal, colloquial, and affective. Each paraphrase must be one or two sentences, name the key edited object, describe the axis-specific change, and use soft preservation language only such as ``broadly similar'' or ``in roughly the same layout''. Do not mention any 3D-FUTURE asset ID, UUID, or coordinate value.

Source room: \texttt{\{spec\}}\\
Operator: \texttt{\{op\}}\\
JSON output: \texttt{\{"formal": "...", "colloquial": "...", "affective": "..."\}}
\end{quote}

\paragraph{LLM Verifier (Section~\ref{sec:method:verifier}).}
\begin{quote}\small\itshape
You are evaluating whether a candidate 3D room satisfies a natural-language edit applied to a reference room. Output a JSON object with two fields: \texttt{confidence}, a continuous value in $[0,1]$ where $1.0$ indicates that the candidate fully satisfies the edit and $0.0$ indicates a clear violation, with intermediate values reflecting mild attribute drift; and \texttt{rationale}, one short sentence. Output JSON only.

Reference room: \texttt{\{spec\_ref\}}\\
Edit instruction: \texttt{\{paraphrase\}}\\
Candidate room: \texttt{\{spec\_cand\}}\\
JSON:
\end{quote}

\section{Implementation Details}
\label{app:implementation}

\paragraph{Mismatch indicator computation.}
The four indicators in Eq.~\ref{eq:costmatrix} are computed by exact string equality on the 3D-FUTURE attribute vocabulary, with the following edge cases. For $\ell_{\mathrm{sty}}$, the room-level fallback compares $s_q$ against the multiset $\{s_{j'}\}_{j' \in C}$ of all object styles in the candidate room and returns $0$ if any $s_{j'}{=}s_q$. For $\ell_{\mathrm{dir}}$, the bounding-box centre coordinate is taken along the axis matching $d_q$. We use the $x$ axis for left and right, the $z$ axis for front and behind, and Euclidean distance to the room centroid for near and far. The indicator is $0$ when the sign of the coordinate matches $d_q$ (negative for ``left'', positive for ``right'', and so on). The non-null rule excludes any term whose corresponding query field is null, so a geometric query with no $s_q$ or $m_q$ is not penalised by missing style or material.

\paragraph{Subject and anchor selection in the structural prior.}
The structural prior identifies a subject and an anchor object inside each candidate. The subject is the first object whose category equals $c_q$. The anchor is the first non-subject object whose category keyword (its first word) appears in the parsed paraphrase. If no such anchor exists, we fall back to the first non-subject object in the candidate's object list. The relative direction is computed as the sign of the bounding-box-centre difference between subject and anchor along the relevant axis, and the bonus rewards candidates whose relative direction agrees with $d_q$. For near and far queries, the bonus thresholds the Euclidean distance between subject and anchor at $1.5\,$m and $3.0\,$m respectively.

\paragraph{Size-class detection.}
For the size-class bonus, we count occurrences of size-keyword tokens in the parsed paraphrase. The vocabulary maps \textit{small / smaller / single / compact / tiny / petite / twin} to the small class, \textit{medium / double / queen} to medium, and \textit{large / larger / king / spacious / big / bigger / huge} to large. The bonus rewards candidates that contain at least one object in the queried category family at the chosen target size class, applies a small penalty when the family is present at the wrong size class, and contributes zero otherwise.

\paragraph{Software.}
We implement the OT backbone using the unbalanced Sinkhorn solver from \textsc{pot} (Python Optimal Transport, $\geq 0.9$). The structural prior is implemented in pure NumPy. The base retrievers are reproduced from the authors' released code where available, and CIReVL and Pic2Word use the same GPT-5.1 model as our parser for fair comparison. All experiments use a single fixed random seed of $42$.

\section{Ablation Heatmap}
\label{app:ablation-heatmap}

\begin{figure}[h]
  \centering
  \includegraphics[width=\linewidth]{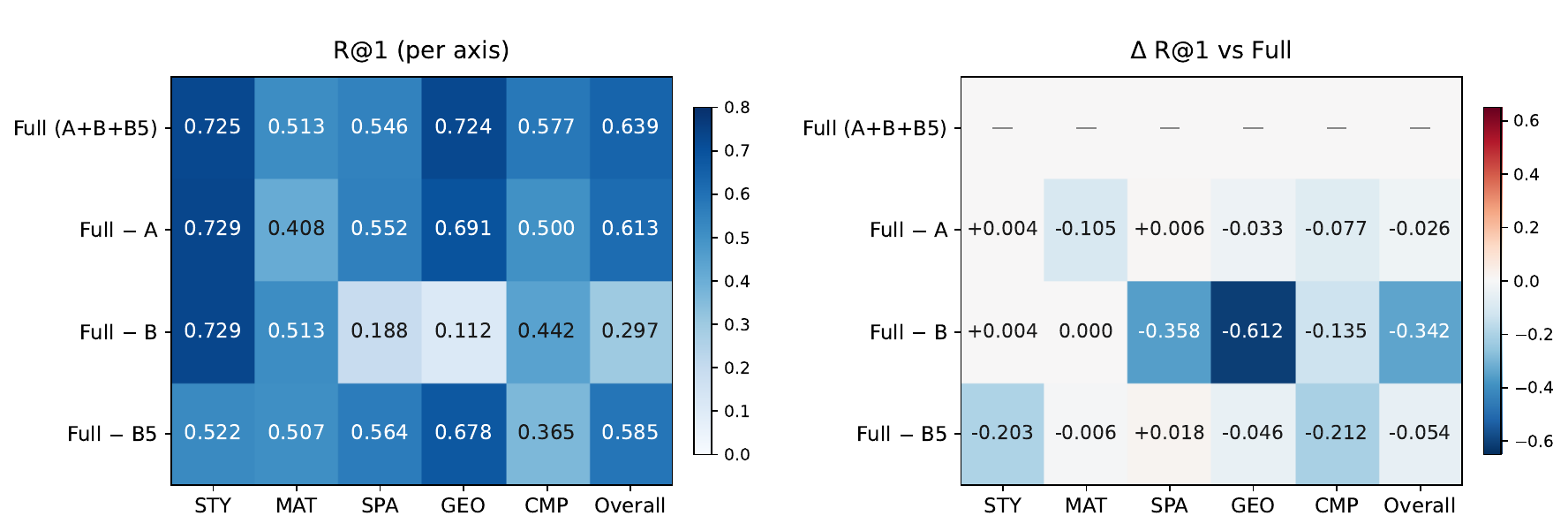}
  \caption{\textbf{Drop-one ablation on the $1{,}274$ with-GT test queries}, visualising Table~\ref{tab:ablation}. Each row removes one component from the full method ($A$: unbalanced OT solver; $B$: structural prior; $B5$: LLM verifier). \textbf{Left}: absolute R@$1$ per edit axis. \textbf{Right}: $\Delta$R@$1$ against the full method, where negative values mean the removed component was needed and deeper blue means a larger drop. Each module's largest negative delta concentrates on its design axis ($A$ on MAT, $B$ on SPA and GEO, $B5$ on STY and CMP), with limited overlap across components.}
  \label{fig:ablation-heatmap}
\end{figure}

Figure~\ref{fig:ablation-heatmap} visualises the drop-one ablation in Table~\ref{tab:ablation} as a per-axis heatmap. The heatmap makes the design-axis correspondence between each component and its target edit type immediately visible. The unbalanced solver's largest drop is on the material axis ($-10.5$ R@$1$), where one-to-many asymmetry is most extreme, since a query about a single object's material has nothing to say about the rest of the room and a balanced solver dilutes the cost across all candidate objects. The structural prior's largest drops are on the geometric ($-61.2$) and spatial ($-35.8$) axes, where bounding-box size cues and pairwise direction cues respectively carry the discriminative signal that the symbolic OT cost matrix cannot capture. The LLM verifier's largest drops are on the style ($-20.3$) and compound ($-21.2$) axes, where holistic linguistic judgment correctly demotes candidates that match keyword-level attributes but fail when read in context.

The off-diagonal pattern of the heatmap is also informative. Each component's largest negative delta concentrates on a different set of axes, with limited overlap. This near-orthogonality of the three components on the per-axis decomposition supports the modular design choice in Section~\ref{sec:method}, where each component handles a structurally distinct signal channel rather than redundantly covering the same axes.

\newpage

\end{document}